\documentclass[10pt,journal,compsoc]{IEEEtran}
\ifCLASSOPTIONcompsoc
  \usepackage[nocompress]{cite}
\else
  \usepackage{cite}
\fi
\ifCLASSINFOpdf
   \usepackage[pdftex]{graphicx}
\else
   \usepackage[dvips]{graphicx}
\fi
\usepackage{amsmath}
\ifCLASSOPTIONcompsoc
 \usepackage[caption=false,font=footnotesize,labelfont=sf,textfont=sf]{subfig}
\else
 \usepackage[caption=false,font=footnotesize]{subfig}
\fi

\usepackage{hyperref}
\usepackage{amsfonts}
\usepackage[T1]{fontenc}
\usepackage{enumerate}

\begin{document}
%
\title{Graph Masked Autoencoders with Transformers}
%
%
%
%

\author{\IEEEauthorblockN{Sixiao Zhang\IEEEauthorrefmark{1},
Hongxu Chen\IEEEauthorrefmark{1}, Haoran Yang,
Xiangguo Sun, Philip S. Yu \IEEEmembership{Fellow, IEEE}, Guandong Xu\IEEEauthorrefmark{2}, \IEEEmembership{Member, IEEE} }

\IEEEcompsocitemizethanks{
\IEEEcompsocthanksitem Sixiao Zhang, Hongxu Chen, Haoran Yang and Guandong Xu are with Data Science and Machine Intelligence Lab, Faculty of Engineering and Information Technology, University of Technology Sydney, Sydney, New South Wales, 2007, Australia.
E-mail: zsx57575@gmail.com, {\{Hongxu.Chen, Guandong.Xu\}}@uts.edu.au, haoran.yang-2@student.uts.edu.au
\IEEEcompsocthanksitem Xiangguo Sun is with the Chinese University of Hong Kong, Hong Kong.
E-mail: xg.sheldon.sun@gmail.com
\IEEEcompsocthanksitem Philip S. Yu is with Department of Computer Science, University of Illinois at Chicago, Chicago, Illinois, 60637, the United States. E-mail: psyu@cs.uic.edu
\protect\\
}
\thanks{* Both authors contributed equally to this research.}
\thanks{\dag\ Corresponding author.}

\thanks{Manuscript received April 19, 2005; revised August 26, 2015.}}

%
%

\markboth{Journal of \LaTeX\ Class Files,~Vol.~14, No.~8, August~2015}%
{Zhang \MakeLowercase{\textit{et al.}}: Graph Masked Autoencoders with Transformers}
%

\IEEEpubid{\makebox[\columnwidth]{\hfill 0000--0000/00/\$00.00~\copyright~2015 IEEE}%
\hspace{\columnsep}\makebox[\columnwidth]{Published by the IEEE Computer Society\hfill}}


\IEEEtitleabstractindextext{%
\begin{abstract}
Recently, transformers have shown promising performance in learning graph representations. However, there are still some challenges when applying transformers to real-world scenarios due to the fact that deep transformers are hard to train from scratch and the quadratic memory consumption w.r.t. the number of nodes. In this paper, we propose Graph Masked Autoencoders (GMAEs), a self-supervised transformer-based model for learning graph representations. To address the above two challenges, we adopt the masking mechanism and the asymmetric encoder-decoder design. Specifically, GMAE takes partially masked graphs as input, and reconstructs the features of the masked nodes. The encoder and decoder are asymmetric, where the encoder is a deep transformer and the decoder is a shallow transformer. The masking mechanism and the asymmetric design make GMAE a memory-efficient model compared with conventional transformers. We show that, when serving as a conventional self-supervised graph representation model, GMAE achieves state-of-the-art performance on both the graph classification task and the node classification task under common downstream evaluation protocols. We also show that, compared with training in an end-to-end manner from scratch, we can achieve comparable performance after pre-training and fine-tuning using GMAE while simplifying the training process.
\end{abstract}

}

\maketitle

\IEEEdisplaynontitleabstractindextext

%
\IEEEpeerreviewmaketitle

\IEEEraisesectionheading{\section{Introduction}\label{sec:introduction}}


%
%
%
%

\IEEEPARstart{A}{s} the Internet and information technology develop rapidly in recent years, graph-structured data has emerged as a new type of information carrier and becomes more and more important in data mining and information retrieval. Graphs can now be seen everywhere in our daily life, such as citation networks \cite{yang2016revisiting, bojchevski2017deep}, social networks \cite{rozemberczki2021multi, zhang2021we}, transportation networks \cite{ribeiro2017struc2vec}, molecule networks \cite{wu2018moleculenet, gomez2018automatic}, and e-commercial networks \cite{zeng2019graphsaint}, etc. They have many important applications in finance, healthcare, recommendation, etc. Different from traditional data formats such as texts and images, graphs do not have a fixed structure and are highly sparse. To overcome the new challenges and to extract useful information from graphs, a lot of machine learning models have been proposed by researchers to model graphs. For example, node2vec \cite{grover2016node2vec}, collaborative filtering \cite{he2017neural}, graph neural networks \cite{kipf2016semi, velickovic2017graph}, etc. 

Graph convolutional neural networks (GCNs) \cite{kipf2016semi} have been a dominating model architecture in graph mining. They extract the local information in graphs by aggregating neighborhood representations. However, it is commonly observed that a 2-layer GCN shows the best overall performance in most cases, while deeper GCNs perform worse \cite{li2018deeper, zhou2020understanding}. That means, GCNs work best when aggregating only 2-hop neighbors. This is believed to be caused by the over-smoothing problem \cite{li2018deeper} introduced by deep GCNs, where the neighborhood aggregating mechanism tends to make all the node representations identical to each other. Although researchers have been attempting to solve this problem in many ways \cite{rong2019dropedge, zhao2019pairnorm, alon2020bottleneck}, such an over-smoothing problem seems to be an inherent limitation of GCNs and is hard to be eliminated. It prevents GCNs from going deeper and thus can only capture local information while ignoring global information. However, such global information has been shown to be also important for learning high-quality graph representations \cite{cao2015grarep, zhu2021pre, buffelli2022impact}. How to effectively capture global information becomes an urgent problem waiting to be solved in the broad graph mining community.

\begin{figure}[t]
\centering
    \includegraphics[width=0.45\textwidth]{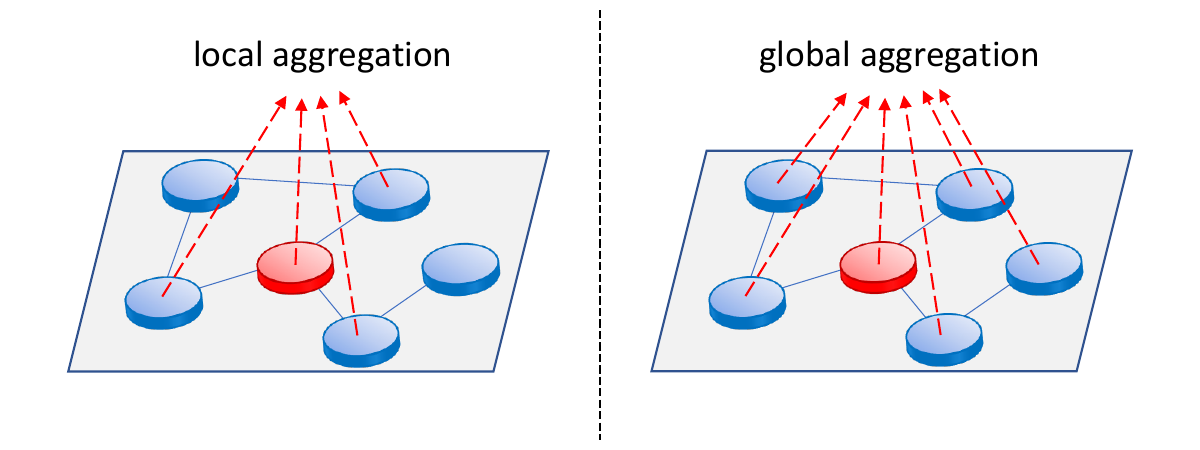}
\caption{An illustration of different aggregation mechanisms used by GCNs and transformers. The red node is the target node. The first picture (left) shows the local aggregation used by a GCN layer, where the embedding of the target node is obtained by aggregating its neighbors. The second picture (right) shows the global aggregation used by a transformer layer, where the embedding is obtained by aggregating all the nodes in the graph.}
\label{fig:intro}
\end{figure}

Recently, transformers \cite{vaswani2017attention} have been applied to the graph domain \cite{ying2021transformers, zhao2021gophormer} and have shown promising performance for solving the above challenge. Transformers are a type of neural network. The core design of them is the self-attention mechanism. In the transformer layer, an attention value is calculated for each instance pair. The embeddings are then calculated by aggregating all other instances weighted by the attentions. An illustration of the difference between GCNs and transformers is shown in \autoref{fig:intro}. Therefore, transformers can easily capture global information. Besides, transformers have a multi-head design to enable effective parallel computing. Because of such good properties, transformers have now become the state-of-the-art model in many NLP and CV tasks \cite{devlin2018bert, dosovitskiy2020image}.




However, there are two key challenges when applying transformers to the graph domain. First, transformers can be very deep with dozens of layers and millions of parameters, which makes them hard to be trained from scratch \cite{zhao2021gophormer}. Tailored training schemes and hyper-parameters are necessary for the model to achieve satisfying performance. Second, transformers have quadratic memory consumption with respect to the number of instances. Real-world graphs usually contain a large number of nodes, which makes running out of memory occur frequently when training transformers on large graphs.

Inspired by BERT \cite{devlin2018bert} and MAE \cite{he2021masked}, we propose a self-supervised graph transformer model named Graph Masked Autoencoders (GMAE). We aim to address the above two challenges with our proposed GMAE. GMAE contains an encoder and a decoder, where the encoder encodes a partially masked graph as low-dimensional embeddings and the decoder reconstructs the features of the masked nodes. To address the challenge of large memory consumption, we use a mask-and-predict mechanism in GMAE, where some of the nodes in the graph are masked, i.e., the number of input nodes is reduced. Since the memory consumption of transformers is quadratic w.r.t. to the number of nodes, the memory consumption of GMAE becomes much less than a conventional transformer. Besides, we use an asymmetric design of encoder and decoder, where the encoder is a deep transformer and the decoder is a shallow transformer (e.g., 2 layers). Such an asymmetric design enables us to train a deeper encoder thanks to the low memory load of the shallow decoder, meanwhile achieving better performance compared with using a deep decoder. To address the other challenge that transformers are hard to train from scratch, we can use GMAE as an effective pre-training tool. After training GMAE in a self-supervised manner, the encoder can be fine-tuned in an end-to-end supervised manner for specific downstream tasks with much less tailored training schemes, while achieving comparable performance with training from scratch. Our implementation is available at \url{https://github.com/RinneSz/GMAE}.

We summarize our contributions as follows:
\begin{itemize}
    \item We propose Graph Masked Autoencoders (GMAEs), a self-supervised model for learning graph representations.
    \item With the masking mechanism and the asymmetric encoder-decoder design, GMAE has a much less memory consumption compared with conventional graph transformers.
    \item GMAE can serve as an effective pre-training tool for graph transformers to simplify the training process.
    \item Extensive experiments show that GMAE achieves state-of-the-art performance on both self-supervised graph classification tasks and self-supervised node classification tasks with common downstream evaluation protocols. GMAE pre-training and fine-tuning scheme also achieves comparable performance with end-to-end training. 
\end{itemize}
\section{Related Work}
\subsection{Self-supervised Graph-level Representation Learning}
Early self-supervised graph-level representation learning methods focus on designing graph kernels. A conventional graph kernel method first decomposes graphs into some subgraphs, then compares the number of shared subgraphs across different graphs to form graph embeddings. Different graph kernel methods usually differ in the way of extracting subgraphs. Popular graph kernel methods include random walk kernels \cite{kashima2003marginalized}, shortest-path kernels \cite{borgwardt2005shortest}, graphlets \cite{prvzulj2007biological, shervashidze2009efficient}, Weisfeiler-Lehman graph kernels \cite{shervashidze2011weisfeiler}, and Deep Graph Kernels \cite{yanardag2015deep}.

Since graph kernels are handcrafted, they suffer from poor generalization ability. To overcome this problem, people have explored many self-supervised contrastive methods to learn graph-level representations. Graph2vec \cite{narayanan2017graph2vec} and sub2vec \cite{adhikari2018sub2vec} use random walk to generate "corpus" of graphs and use skipgram model to learn representations. Infograph \cite{sun2019infograph} maximizes the mutual information between the graph-level representation and the representations of substructures. Recently, graph contrastive learning has emerged as the new state-of-the-art self-supervised graph representation learning method, which maximizes the agreement between two different views of the original graph. Typical graph-level contrastive learning methods include GraphCL \cite{you2020graph}, MVGRL \cite{hassani2020contrastive}, GCC \cite{qiu2020gcc}, etc.

\subsection{Self-supervised Node-level Representation Learning}
Traditional self-supervised node-level representation learning methods, such as DeepWalk \cite{perozzi2014deepwalk} and node2vec \cite{grover2016node2vec}, adopt random walks to extract local structures and force nodes that are close to each other to have similar embeddings. This is a strict assumption that may not always be true. Besides, such methods focus on the structural information of graphs and ignore the feature information.

As graph neural networks develop rapidly in recent years, successful attempts have been made in applying GNNs to self-supervised node-level representation learning. For example, VGAE/GAE \cite{kipf2016variational} and GraphSAGE \cite{hamilton2017inductive}, where VGAE/GAE reconstructs the adjacency matrix and GraphSAGE forces the nearby nodes to have similar representations. Most recently, a large number of works that apply graph contrastive learning to the node-level representation learning scenario have emerged and shown state-of-the-art performance. Similar to graph contrastive learning methods applied in the graph-level representation learning scenario, these works in the node-level representation learning scenario also differ in the way they generate different views of the graph. Representative methods include DGI \cite{velickovic2019deep}, GRACE \cite{zhu2020deep}, and GCA \cite{zhu2021graph}.

\subsection{Transformers}
Transformers \cite{vaswani2017attention} are one type of neural network which uses the attention mechanism as the core design. It has achieved state-of-the-art performance in many tasks in both natural language processing (NLP) and computer vision (CV). In NLP, a large number of transformer models have been proposed to solve various tasks. For example, MCA \cite{liu2018generating}, sparse transformers \cite{child2019generating}, Longformer \cite{beltagy2020longformer}, BigBird \cite{zaheer2020big}, Reformer \cite{kitaev2020reformer}, Linear Transformer \cite{katharopoulos2020transformers}, etc. Among such a great number of works, BERT \cite{devlin2018bert} should be the most well-known language model built with transformers.

While transformers have become the dominating model in NLP, people have also tried to apply transformers in the computer vision domain. They have again achieved state-of-the-art performance in many different vision tasks, such as image classification \cite{dosovitskiy2020image, touvron2021training, radford2021learning}, object detection \cite{carion2020end}, and few-shot learning \cite{doersch2020crosstransformers}, etc.


Early applications of transformers in graph domain focus on the sequential recommendation because the positional embeddings are easy to be defined based on the user interaction sequence and timestamps, such as SASRec \cite{kang2018self}, BERT4Rec \cite{sun2019bert4rec}, SSE-PT \cite{wu2020sse}, TGSRec \cite{fan2021continuous}, etc. Some other works have tried to adopt transformers into static graph tasks and have achieved competitive performance compared with state-of-the-art GNN-based models. For example, GROVER \cite{rong2020self} incorporates traditional GNNs into the transformer architecture to help process graph-structured data. GT \cite{dwivedi2020generalization} uses the Laplacian eigenvector as the positional embeddings. MAT \cite{maziarka2020molecule} achieves competitive performance on molecular prediction tasks by altering the attention mechanism in transformers with inter-atomic distances and the molecular graph structure. SAN \cite{kreuzer2021rethinking} uses a full Laplacian spectrum to learn the positional embeddings of nodes. Graphormer \cite{ying2021transformers} uses degree centralities, shortest path distances, and edge features as the positional embeddings. Gophormer \cite{zhao2021gophormer} extends graph transformers into node-level tasks for a single large graph by using the ego-graphs instead of the full-graphs. These works are all done in an end-to-end supervised manner. Most of them are hard to be trained from scratch \cite{zhao2021gophormer} and suffer from the large quadratic memory consumption with respect to the input length.
\section{Preliminaries}
\subsection{Notations and Definitions}
A graph $G$ can be described as $G=(V,E)$, where $V$ is the set of vertices (nodes) and $E$ is the set of edges. Suppose there are $n_{V}$ nodes and $n_{E}$ edges in a graph, an adjacency matrix $A\in \mathbb{R}^{n_{V}\times n_{V}}$ is used to describe the connections among nodes. Each entry in $A$ is either 1 or 0 representing whether there is an edge between the two nodes or not. Some graphs have auxiliary information such as node features $X_{V}\in \mathbb{R}^{n_{V}\times d_{V}}$ and edge features $X_{E}\in \mathbb{R}^{n_{E}\times d_{E}}$ which can be used to facilitate training, where $d_{V}$ and $d_{E}$ are the numbers of dimensions of node features and edge features respectively.

Graph self-supervised learning aims to train a model $f$ that takes the graph data (without labels) as input and outputs the embeddings of each node. In this work, we assume that we only have node features and edge features as auxiliary information. Thus, the model $f$ can be expressed as
\begin{equation}
    f(A,X_{V},X_{E})=H_{o}
\end{equation}
where $H_{o}\in\mathbb{R}^{n_{V}\times d_{o}}$ contains the output embeddings of each node, and $d_{o}$ is the number of output dimensions.

\subsection{Graph Transformers}
\label{pre:graph transformer}
A transformer is composed of several transformer layers. Each transformer layer contains a self-attention module and a feed-forward network. Suppose the input hidden representation matrix is $H\in\mathbb{R}^{n\times d}$, where $n$ is number of instances (nodes) and $d$ is the number of hidden dimensions. The self-attention module contains three learnable matrices $W_{Q}\in\mathbb{R}^{d\times d_{q}}$, $W_{K}\in\mathbb{R}^{d\times d_{k}}$ and $W_{V}\in\mathbb{R}^{d\times d_{v}}$, which stand for \textbf{Queries}, \textbf{Keys} and \textbf{Values} respectively. Note that we usually have $d_{q}=d_{k}$ due to the self-attention mechanism introduced next. The self-attention module first computes the three matrices as
\begin{equation}
    Q=HW_{Q},\ K=HW_{K},\ V=HW_{V}
\end{equation}
Then the dot product is applied to the queries with all keys to obtain the weights on the values. The final output matrix is computed by
\begin{equation}
    \text{Attention}(Q,K,V)=\text{softmax}(\frac{QK^{T}}{\sqrt{d_{k}}}V)
\end{equation}
The obtained output matrix is a matrix with size $n\times d_{v}$ where each row represents the output representation of the corresponding instance.

Then a 2-layer feed-forward network (FFN) is used to project the node representations $x$ and increase model capacity:
\begin{equation}
    FFN(x)=\max(0,xW_{1}+b_{1})W_{2}+b_{2}
\end{equation}
where $W_{1}$ and $W_{2}$ are learnable weight matrices. $b_{1}$ and $b_{2}$ are learnable biases.

The node feature information is already contained in the input feature matrix and is seen by the transformer. But the aforementioned forward propagation does not contain the structural information of graphs, i.e., each node is treated identically. Therefore, positional embeddings are necessary for the transformer to know the relative positions of nodes. In natural language processing and computer vision, the positional embeddings are easy to define based on the absolute positions of words and pixels. However, unlike sentences and images, nodes in a graph do not have an absolute order. We don't know which node is the "first" node in a graph. Nodes can only be described with relative positions based on their connections. To solve this problem, Graphormer \cite{ying2021transformers} proposed three types of positional encodings to jointly capture the structural information of graphs, which are the centrality encoding, the spatial encoding, and the edge encoding. They encode the degree centralities, the shortest path distances, and the edge features respectively. In this way, the graph structural information can be well captured by the transformer.



\section{Graph Masked Autoencoder}
\label{sec:methodology}

\begin{figure}[t]
\centering
    \includegraphics[width=0.45\textwidth]{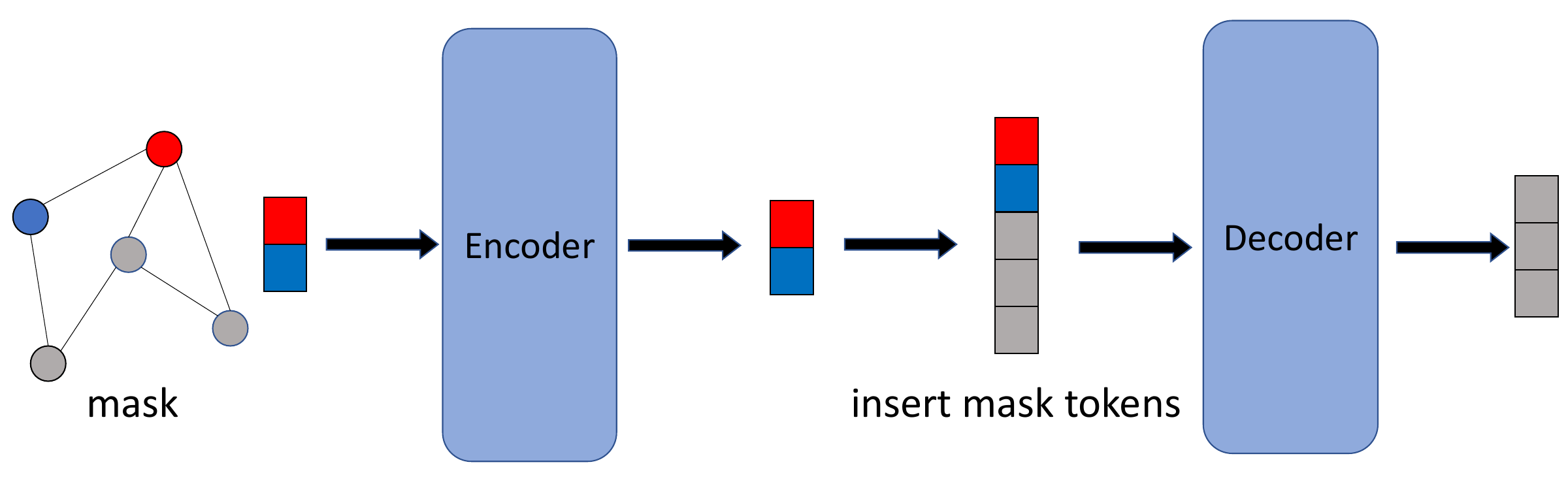}
\caption{GMAE framework. We first randomly mask some of the nodes (the grey nodes). The feature matrix and positional embeddings of the non-masked nodes (the red node and the blue node) are fed into an encoder to obtain the representations of the non-masked nodes. Then we insert a shared learnable mask token to represent the masked nodes. The decoder reconstructs the original features of the masked nodes.}
\label{fig:GMAE}
\end{figure}

In this section, we introduce the details of our proposed graph masked autoencoders (GMAEs). A GMAE contains an encoder and a decoder. We use Graphormer \cite{ying2021transformers} as our backbone model, i.e., our encoder and decoder are both graph transformers proposed by the Graphormer.

\subsection{Training}
We first introduce how GMAE works by going through the training process. An illustration of the GMAE framework is shown in \autoref{fig:GMAE}. Note that we omit how we deal with the positional embeddings here and it will be introduced afterward.

Generally, the forward propagation of GMAE contains four steps:
\begin{enumerate}[i)]
    \item Randomly mask nodes in the input graph.
    \item Feed the non-masked nodes into the encoder and obtain the embeddings of them.
    \item Use a shared learnable mask token to represent the embeddings of the masked nodes and insert them into the output of the encoder.
    \item Feed the embedding matrix with inserted mask tokens into the decoder to reconstruct the features of the masked nodes.
\end{enumerate}

Now we explain it in detail. During training, for each input graph, we first randomly mask some of the nodes. These masked nodes will not be seen by the encoder. The encoder will only take the features of the observed nodes as input and then outputs the embeddings of each observed node. For example, suppose the number of nodes in an input graph is $n$ and the number of node features is $d_{V}$. Conventional end-to-end graph transformers take the full feature matrix $X_{V}\in\mathbb{R}^{n\times d_{V}}$ as input. However, the encoder in GMAE only takes partial feature matrix $X_{V}^{o}\in\mathbb{R}^{n_{o}\times d_{V}}$ as input, where $n_{o}$ is the number of observed nodes that are not masked. The output embedding matrix of the encoder is $X_{e}\in\mathbb{R}^{n_{o}\times d_{e}}$, where $d_{e}$ is the number of output dimensions. Each row in $X_{e}$ is the embedding of the corresponding observed node.

The decoder aims to reconstruct the masked node features given the observed node embeddings. After we obtain the output embedding matrix $X_{e}$ from the encoder, we randomly initialize a mask token $x_{m}\in\mathbb{R}^{1\times d_{e}}$ to replace the masked nodes and insert them into $X_{e}$ to obtain a new matrix $X_{e}'\in\mathbb{R}^{n\times d_{e}}$. The mask token $x_{m}$ is shared across all masked nodes, i.e., all the masked nodes use the same mask token. The mask token $x_{m}$ will be updated in each iteration through back-propagation just like all other learnable parameters. The output of the decoder is the reconstructed node features of the masked nodes. A typical reconstruction loss such as the cross-entropy loss is applied to the reconstructed features and the original features of the masked nodes.

\subsection{Positional Embeddings}
We follow Graphormer and use their proposed centrality encoding, spatial encoding, and edge encoding as the positional embeddings. To obtain the encodings, we need to know the node degrees, the shortest path distances of all node pairs, and the edge features. Note that edge encoding only works on datasets where edge features are available. 

In GMAE, we obtain the positional embeddings of nodes based on the whole graph, i.e., our masking mechanism is only applied to the node features and not to the graph structure. When calculating the node degrees and shortest path distances, the whole graph is presented to us without masking. Then the positional embeddings will be added to the input of the encoder and the decoder, and also to the attention matrix in each transformer layer. Please refer to Graphormer \cite{ying2021transformers} for more details regarding the positional embeddings.


\subsection{Evaluation}
After training, the decoder is discarded, and we only use the encoder for downstream tasks. At the inference stage, no masking is applied, and the whole graph is fed into the encoder. For graph-level tasks, a readout function can be applied to the node embeddings, such as the average pooling. These node embeddings and graph embeddings can be further used for downstream tasks such as classification and clustering. GMAE can be used for both vanilla self-supervised tasks and pre-training tasks. For pre-training tasks, the encoder will be fine-tuned according to the specific requirements of the tasks. In self-supervised tasks, we can simply fix the parameters of the encoder, and the output node embeddings and graph embeddings are used as the input for downstream models and tasks.

\subsection{Properties}
GMAE has a much less memory consumption compared with conventional end-to-end graph transformers because of the masking mechanism and the asymmetric encoder-decoder design. They both help save a lot of memory and make it possible for GMAE to be trained on larger graphs and to train deeper graph transformer encoders. 
\subsubsection{Masking mechanism} 
Because of the masking mechanism, the size of the input feature matrix is largely reduced. We will see in our experiments that masking half of the nodes yields good overall performance on most of the datasets, while for some particular datasets, the best mask ratio can even reach 90\%. Since graph transformers have quadratic memory consumption with respect to the input length, such a high mask ratio can largely reduce the memory consumption.
\subsubsection{Asymmetric encoder-decoder design}
In GMAE, the encoder and the decoder are asymmetric. The encoder is a deep graph transformer while the decoder is a shallow graph transformer. For example, in our experiments, an encoder with 16 layers and a decoder with 2 layers can achieve state-of-the-art performance in most cases. Such a design leads to an expressive encoder and meanwhile saves computation resources. Although our encoder is deep, the size of the input feature matrix is reduced because of the masking mechanism. However, a conventional end-to-end graph transformer is a deep transformer just like our encoder, but with a full feature matrix as the input, which makes the memory consumption quite large. On the other hand, the input to our decoder is an embedding matrix of full size, which seems to have a large memory consumption. In fact, since the decoder is shallow, the computational load is still quite small.
\subsubsection{Scalability}
Because of the above two designs, GMAE can be potentially used to train large graphs and deep encoders. 
\begin{itemize}
    \item \textbf{Scalable to large graphs.} By masking half or more nodes, GMAE can be used to learn graphs with a large size while still maintaining a relatively small memory consumption.
    \item \textbf{Scalable to deep encoders.} Since the input to the encoder is a masked feature matrix, increasing the number of layers in the encoder will not cause the memory consumption to increase as much as a conventional transformer that takes the full feature matrix as input. Therefore, we can potentially train a deeper encoder which can increase the model expressiveness.
\end{itemize}


\section{Experiments}
We show the performance of GMAE by three groups of experiments. We first show how GMAE performs as a vanilla self-supervised model by comparing it with popular self-supervised graph representation models on both the graph classification task and the node classification task. Then we show how GMAE performs with pre-training and fine-tuning by comparing it with state-of-the-art supervised models on the graph classification task.

\begin{table}[!t]
\renewcommand\arraystretch{1.3}
\caption{Dataset Statistics for Graph Classification}
\label{table:datasets}
\centering
 \begin{tabular}{c||c|c|c} 
 \hline
  & \# Graphs & Avg. Nodes & Avg. Edges\\
 \hline
 \hline
  PROTEINS & 1113 & 39.06 & 72.82\\
 \hline
 BZR & 405 & 35.75 & 38.36\\
 \hline
 COX2 & 467 & 41.22 & 43.45\\
 \hline
 NCI1 & 4110 & 29.87 & 32.30\\
 \hline
 MUTAG & 188 & 17.93 & 19.79\\
 \hline
 ER\_MD & 446 & 21.33 & 234.85\\
 \hline
 DHFR & 467 & 42.43 & 44.54\\
 \hline
\end{tabular}
\end{table}

\begin{table*}[!t]
\renewcommand\arraystretch{1.3}
\caption{Self-supervised Graph Classification Accuracy in Percentage}
\label{table:graph classification performance}
\begin{center}
 \begin{tabular}{c||c|c|c|c|c|c|c} 
 \hline
  & PROTEINS & BZR & COX2 & NCI1 & MUTAG & ER\_MD & DHFR\\
 \hline
 \hline
 sub2vec & 67.31$\pm$0.50 & 78.52$\pm$0.50 & 78.07$\pm$0.17 & 59.76$\pm$0.35 & 84.47$\pm$0.37 & 59.41$\pm$0.00 & 60.95$\pm$0.21\\
 \hline
 graph2vec & 67.31$\pm$0.41 & 84.00$\pm$1.08 & 80.15$\pm$0.58 & 69.66$\pm$0.26 & 86.44$\pm$1.25 & 65.82$\pm$1.52 & 77.29$\pm$0.87 \\
 \hline
 GCN-SAGE & 59.57$\pm$0.00 & 78.77$\pm$0.00 & 78.16$\pm$0.00 & 50.05$\pm$0.00 & 66.49$\pm$0.33 & 59.41$\pm$0.00 & 60.98$\pm$0.00 \\
 \hline
 Graphormer-SAGE & 72.13$\pm$0.55 & 83.87$\pm$0.25 & 78.21$\pm$0.37 & 69.40$\pm$0.42 & 87.21$\pm$0.56 & 68.43$\pm$1.76 & 60.98$\pm$0.51 \\
 \hline
 InfoGraph & 74.02$\pm$0.40 & \underline{84.84$\pm$0.86} & 80.55$\pm$0.51 & 77.50$\pm$0.74 & 86.07$\pm$1.78 & \underline{72.24$\pm$0.88} & \textbf{80.48$\pm$1.34} \\
 \hline
 GraphCL & \underline{74.89$\pm$0.65} & 84.20$\pm$1.82 & \underline{81.10$\pm$0.82} & \underline{78.75$\pm$0.28} & \underline{87.66$\pm$1.03} & \textbf{73.23$\pm$0.86} & 68.81$\pm$4.15 \\
 \hline
 \textbf{GMAE} & \textbf{75.38$\pm$0.85} & \textbf{87.43$\pm$0.99} & \textbf{81.75$\pm$0.79} & \textbf{82.51$\pm$0.31} & \textbf{88.97$\pm$0.83} & 71.14$\pm$0.74 & \underline{80.06$\pm$1.73}
 \\
 \hline
\end{tabular}
\end{center}
\end{table*}

\subsection{GMAE Self-supervised Training for Graph Classification}
We show that GMAE has state-of-the-art performance as a vanilla self-supervised method for the graph classification task by comparing it with popular self-supervised graph representation baselines under the commonly used SVM evaluation protocol, where we use a downstream SVM classifier to conduct the graph classification task to evaluate the quality of the learned graph embeddings. 

\subsubsection{Baselines}
We choose 6 self-supervised graph representation models as our baselines:
\begin{itemize}
    \item \textbf{sub2vec} \cite{adhikari2018sub2vec}: learning representations for subgraphs using random walks.
    \item \textbf{graph2vec} \cite{narayanan2017graph2vec}: graph-level representation learning using random walks.
    \item \textbf{GCN} \cite{kipf2016semi}: graph convolutional networks.
    \item \textbf{Graphormer} \cite{ying2021transformers}: graph-level transformers. 
    \item \textbf{InfoGraph} \cite{sun2019infograph}:  mutual information maximization between graph-level representation and the representations of substructures of different scales.
    \item \textbf{GraphCL} \cite{you2020graph}: a graph-level contrastive learning framework.
\end{itemize}
Sub2vec and graph2vec are representative self-supervised graph representation models based on random walks. InfoGraph and GraphCL are the state-of-the-art self-supervised graph contrastive learning models. However, as for GCN and Graphormer, since they are originally semi-supervised and supervised methods respectively, we adopt the graph loss introduced in GraphSage \cite{hamilton2017inductive} to them to enable self-supervised representation learning.

\subsubsection{Datasets}
We use 7 publicly available benchmark graph classification datasets, namely PROTEINS, BZR, COX2, NCI1, MUTAG, ER\_MD, and DHFR. They are collected by the TU Dortmund University \cite{KKMMN2016}.  Dataset statistics are listed in \autoref{table:datasets}. We use 10-fold cross-validation for all the datasets.

\subsubsection{Experiment Settings}
For all the baseline methods, we follow their codes and settings and report the best results. For GMAE, the number of encoder layers is tuned from 1 to 30, and the default number of decoder layers is 2, while a sensitivity analysis on the number of decoder layers is provided in the analysis section. The mask ratio is tuned from 0.1 to 0.9 with step size 0.1. The number of hidden dimensions is 80 and the number of attention heads in the transformer layers is 8. We use a linear decay learning rate scheduler with a 40k-step warm-up stage and 400k maximum training steps. We set the peak learning rate to be 1e-4 and the end learning rate to be 1e-9. The training stops when the loss does not decrease for 50 epochs. After obtaining the learned graph embeddings, we use an SVM classifier to do graph classification with 10-fold cross-validation following \cite{sun2019infograph, you2020graph}. We report the average accuracy of 5 runs.

\subsubsection{Results}
\autoref{table:graph classification performance} shows the results of all methods. The boldfaced ones are the best, and the underlined ones are the second. GMAE shows state-of-the-art performance by outperforming other methods on 5 of the 7 datasets. We can further observe that, among the 6 baseline methods, graph contrastive learning methods (InfoGraph and GraphCL) achieve the best performance. The self-supervised Graphormer also performs well but is a little bit worse than InfoGraph and GraphCL. Instead, our GMAE outperforms InfoGraph and GraphCL on 5 of the 7 datasets. This suggests that our GMAE framework can effectively boost the performance of graph transformers in the self-supervised graph classification scenario.

\begin{table}[!t]
\renewcommand\arraystretch{1.3}
\caption{Dataset Statistics for Node Classification}
\label{table:node datasets}
\centering
 \begin{tabular}{c||c|c|c|c} 
 \hline
  & \# Nodes & \# Edges & \# Features & \# Classes \\
 \hline
 \hline
  Cora & 2708 & 5429 & 1433 & 7\\
 \hline
 CiteSeer & 3327 & 4552 & 3703 & 6\\
 \hline
 PubMed & 19717 & 44324 & 500 & 3\\
 \hline
 Wiki-CS & 11701 & 216123 & 300 & 10\\
 \hline
 Computers & 13752 & 245861 & 767 & 10\\
 \hline
 Photo & 7650 & 119081 & 745 & 8\\
 \hline
\end{tabular}
\end{table}

\begin{table*}[!t]
\renewcommand\arraystretch{1.3}
\caption{Self-supervised Node Classification Accuracy in Percentage}
\label{table:node classification performance}
\begin{center}
 \begin{tabular}{c||c|c|c|c|c|c} 
 \hline
  & Cora & CiteSeer & PubMed & Wiki-CS & Computers & Photo \\
 \hline
 \hline
 GAE & 80.11$\pm$0.22 & 65.75$\pm$0.30 & 77.50$\pm$0.15 & 76.43$\pm$0.07 & 86.66$\pm$0.07 & 91.85$\pm$0.06 \\
 \hline
 DGI & 80.52$\pm$0.15 & \underline{71.40$\pm$0.15} & 76.66$\pm$0.18 & 75.35$\pm$0.14 & 83.95$\pm$0.47 & 91.61$\pm$0.22 \\
 \hline
 MVGRL & \underline{80.86$\pm$0.25} & \textbf{73.00$\pm$0.00} & 79.65$\pm$0.10 & \underline{77.77$\pm$0.06} & \underline{87.39$\pm$0.11} & 91.74$\pm$0.07 \\
 \hline
 GCA & 78.22$\pm$0.58 & 67.52$\pm$1.30 & \underline{81.28$\pm$0.85} & 77.30$\pm$0.83 & 87.35$\pm$0.35 & \underline{92.28$\pm$0.41} \\
 \hline
 \textbf{GMAE} & \textbf{81.14$\pm$0.72} & 69.25$\pm$1.53 & \textbf{81.40$\pm$0.81} & \textbf{78.47$\pm$0.42} & \textbf{87.46$\pm$0.22} & \textbf{93.59$\pm$0.30} \\
 \hline
\end{tabular}
\end{center}
\end{table*}

\subsection{GMAE Self-supervised Training for Node Classification}
We further show how GMAE performs on the self-supervised node classification task. We compare it with several state-of-the-art self-supervised models under the logistic regression evaluation protocol, where we use a downstream logistic regression model to conduct the node classification task to evaluate the quality of the node embeddings.

\subsubsection{Baselines}
We compare GMAE with 4 popular self-supervised node classification models. Specifically, they include:
\begin{itemize}
    \item \textbf{GAE} \cite{kipf2016variational}: the graph autoencoder, a GCN-based autoencoder that reconstructs the adjacency matrix.
    \item \textbf{DGI} \cite{velickovic2019deep}: deep graph infomax, a graph contrastive learning model based on maximizing the mutual information between different representations.
    \item \textbf{MVGRL} \cite{hassani2020contrastive}: a graph contrastive learning method that contrasting encodings from first-order neighbors and a graph diffusion.
    \item \textbf{GCA} \cite{zhu2021graph}: graph contrastive learning with different centrality-based augmentation schemes such as the degree centrality and the eigenvector centrality. 
\end{itemize}
These baselines are now the state-of-the-arts in self-supervised node classification tasks. They are all representative models that are built based on the graph neural networks and outperform traditional self-supervised methods such as DeepWalk \cite{perozzi2014deepwalk} and node2vec \cite{grover2016node2vec}.

\subsubsection{Datasets}
We use 6 publicly available benchmark node classification datasets. They include three citation networks Cora, CiteSeer, and PubMed introduced in \cite{yang2016revisiting}; a Wikipedia-based dataset Wiki-CS from \cite{mernyei2020wiki}; and the Amazon-Computers and Amazon-Photo datasets from \cite{shchur2018pitfalls}. The statistics of the datasets are shown in \autoref{table:node datasets}. For the train/val/test split, we use the public split for Cora, CiteSeer, PubMed, and Wiki-CS. For Amazon-Computers and Amazon-Photo, we randomly split the datasets into 80\%/10\%/10\% train/val/test split.

\subsubsection{Experiment Settings}
For all the 4 baseline models, we follow their implementations and report the best results we got. For GMAE, we follow Gophormer \cite{zhao2021gophormer} to extract ego-graphs for each node to train the model. Specifically, we adopt the neighbor sampler as introduced in GraphSAGE \cite{hamilton2017inductive}, which generates subgraphs by randomly sampling a certain number of nodes in the neighborhood of the target node. These subgraphs (ego-graphs) are then used to train the GMAE model. However, we do not use the proximity positional encodings as introduced in Gophormer \cite{zhao2021gophormer}. Instead, we keep using the positional encodings introduced in Graphormer \cite{ying2021transformers} because we found that the node classification result is better when using the positional encodings in Graphormer.

In each training epoch, we randomly sample one ego-graph for each node and use them to train the GMAE model. We set the depth of the ego-graphs to be 2, i.e., the nodes in the ego-graphs are within the 2-hop neighborhood. The number of neighbors to sample for each node is tuned from 1 to 10. For each ego-graph, we randomly mask a certain portion of nodes according to the mask ratio, and reconstruct the features of the masked nodes. At the inference stage, for each target node, we randomly sample 10 ego-graphs and feed them to the model to obtain embeddings of each node. The embedding of the target node is calculated by averaging the 10 embeddings of the target node in the 10 ego-graphs. The way we do hyper-parameter tuning is the same as what we did in the previous experiment for self-supervised graph classification. After we obtain the embeddings of each node, we use them to train a logistic regression model to conduct node classification. We report the average node classification accuracy of 5 runs.

\subsubsection{Results}
\autoref{table:node classification performance} shows the experiment results of all the methods. The boldfaced ones are the best, and the underlined ones are the second. GMAE shows promising performance by outperforming the four baselines on 5 of the 6 datasets, including the state-of-the-art graph contrastive learning models such as DGI, MVGRL, and GCA. We can conclude that GMAE is an effective self-supervised model at both the graph level and the node level. The representations learned by GMAE can be used for various downstream tasks.

\begin{table}[t]
\renewcommand\arraystretch{1.3}
\caption{GMAE Pre-training and Fine-tuning Result on ZINC}
\label{table:ZINC}
\centering
 \begin{tabular}{c|c} 
 \hline
  Method & Test MAE \\
  \hline
 \hline
 GIN & 0.526$\pm$0.051\\
 \hline
 GraphSage & 0.398$\pm$0.002\\
 \hline
 GAT & 0.384$\pm$0.007\\
 \hline
 GCN & 0.367$\pm$0.011\\
 \hline
 GatedGCN-PE & 0.214$\pm$0.006\\
 \hline
 MPNN(sum) & 0.145$\pm$0.007\\
 \hline
 PNA & 0.142$\pm$0.010\\
 \hline
 GT & 0.226$\pm$0.014\\
 \hline
 SAN & 0.139$\pm$0.006\\
 \hline
 Graphormer & 0.122$\pm$0.006\\
 \hline
  \textbf{GMAE} & \textbf{0.120$\pm$0.003}\\
  \hline
\end{tabular}
\end{table}

\subsection{GMAE Pre-training \& Fine-tuning}
We then compare GMAE pre-training and fine-tuning with state-of-the-art supervised models. We conduct experiments on the ZINC subset dataset \cite{dwivedi2020benchmarking}, which contains 10000 molecular graphs for training, 1000 for validation, and 1000 for testing. The task is to regress the penalized logP value for each graph, which is defined by $y=\text{logP}-\text{SAS}-\text{cycles}$, where $\text{logP}$ is the water-octanol partition coefficient, $\text{SAS}$ is the synthetic accessibility score, and $\text{cycles}$ is the number of cycles with more than six atoms. Graphormer \cite{ying2021transformers} has shown state-of-the-art performance on ZINC, where the model they used is a 12-layer graph transformer. For a fair comparison, in our GMAE pre-training and fine-tuning, we also use a 12-layer graph transformer as the encoder. We use a 2-layer graph transformer as the decoder in the pre-training phase. During pre-training, we train the encoder and the decoder in a self-supervised manner as introduced in \autoref{sec:methodology}. Only the training set is used to pre-train the model. The hyper-parameters used in the pre-training phase are consistent with Graphormer, except that we stop early when the loss does not decrease for a certain number of epochs. Specifically, the number of hidden dimensions is 80, the batch size is 256, the number of attention heads in the transformer layer is 8, the dropout rate is 0.1, and the peak learning rate is 1e-4 with a 40k-step warm-up stage, followed by a linear decay learning rate scheduler where the end learning rate is 1e-9. The maximum number of training steps is 400k. In the fine-tuning phase, we discard the decoder, then fine-tune the encoder and an additional linear projection layer in a supervised manner using the training set. We fine-tune it for 600 epochs with a 1e-3 learning rate. We report the mean absolute error (MAE) on the test set. A smaller MAE indicates a better performance. The final result is shown in \autoref{table:ZINC}. The compared numbers are from the Graphormer paper \cite{ying2021transformers}.

\begin{figure}[!t]
\centering
    \includegraphics[width=0.4\textwidth]{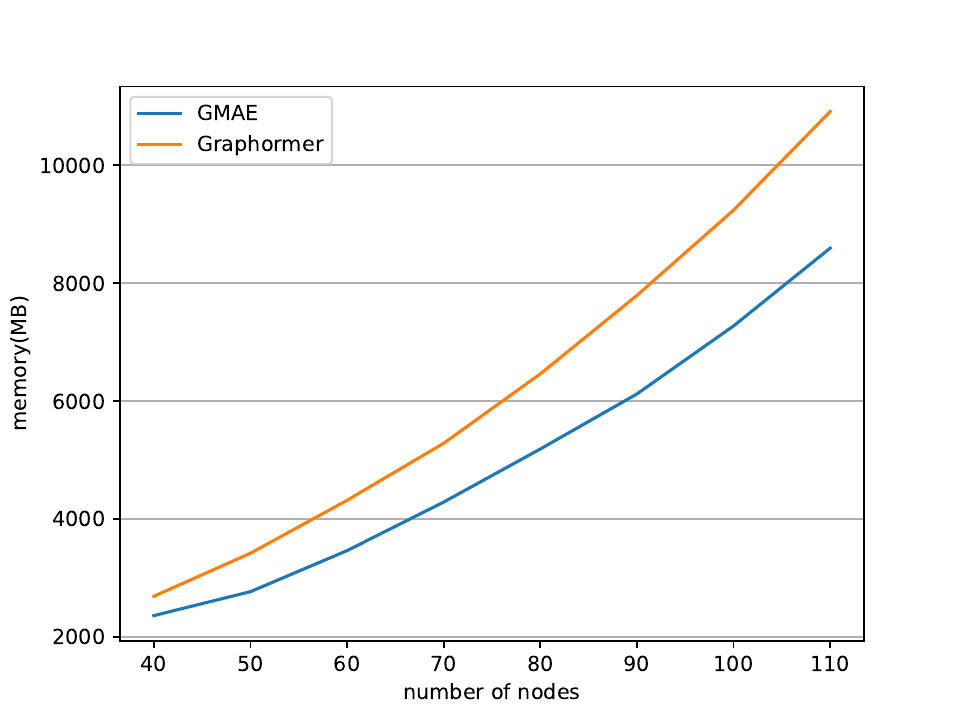}
\caption{Memory usage of GMAE and Graphormer with different number of nodes.}
\label{fig:memory}
\end{figure}

\begin{figure*}[t]
\centering
    \subfloat[PROTEINS]{\includegraphics[width=0.25\linewidth]{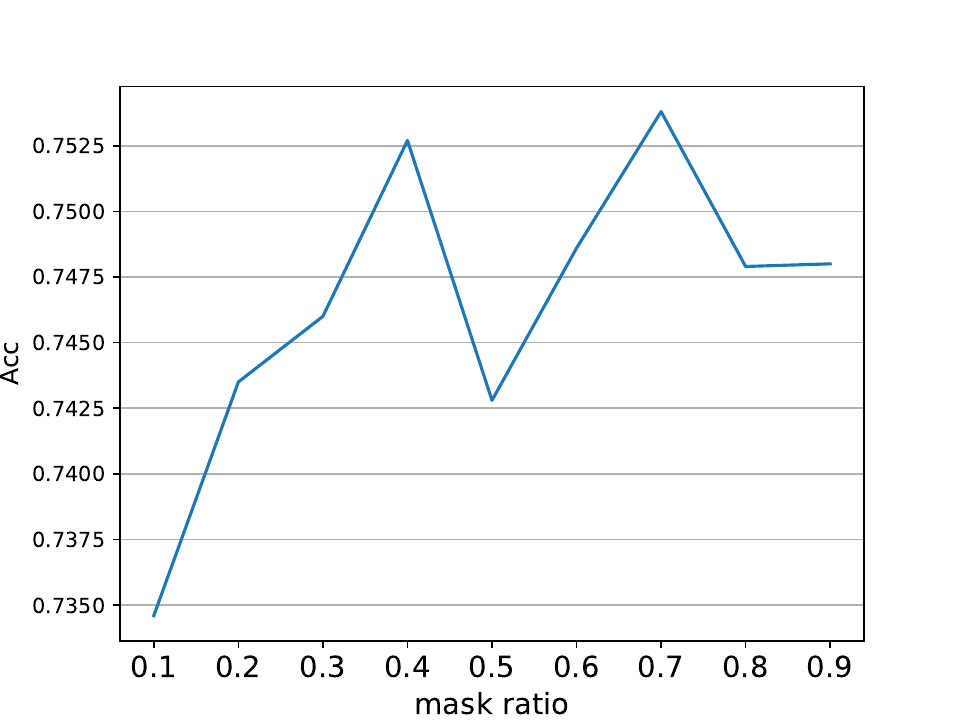}\label{fig:proteins mask ratio}}
    \subfloat[BZR]{\includegraphics[width=0.25\linewidth]{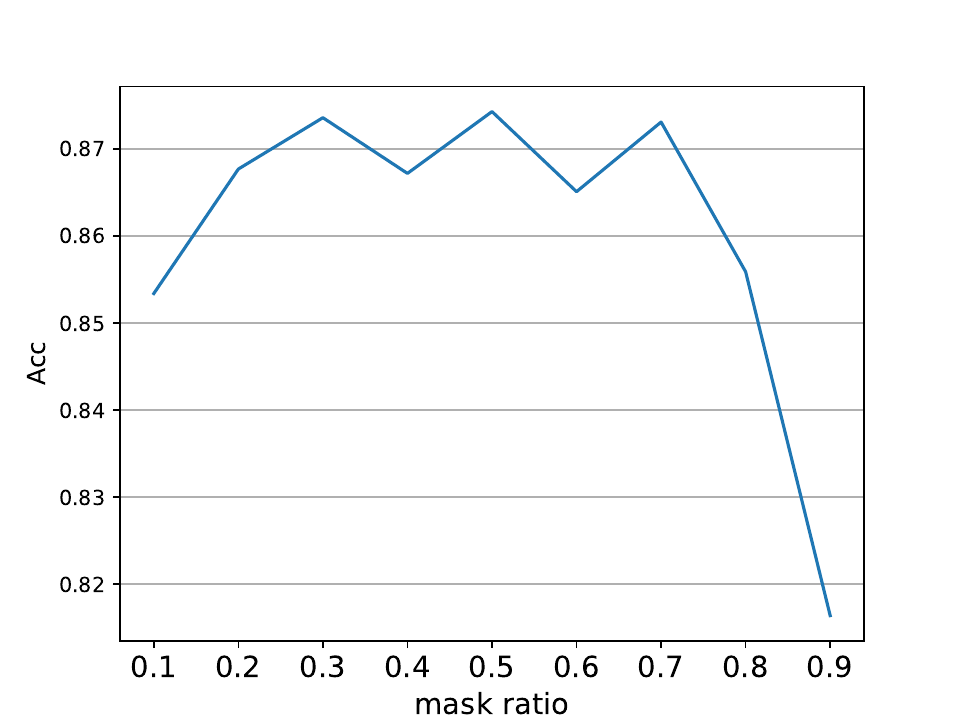}\label{fig:bzr mask ratio}}
    \subfloat[COX2]{\includegraphics[width=0.25\linewidth]{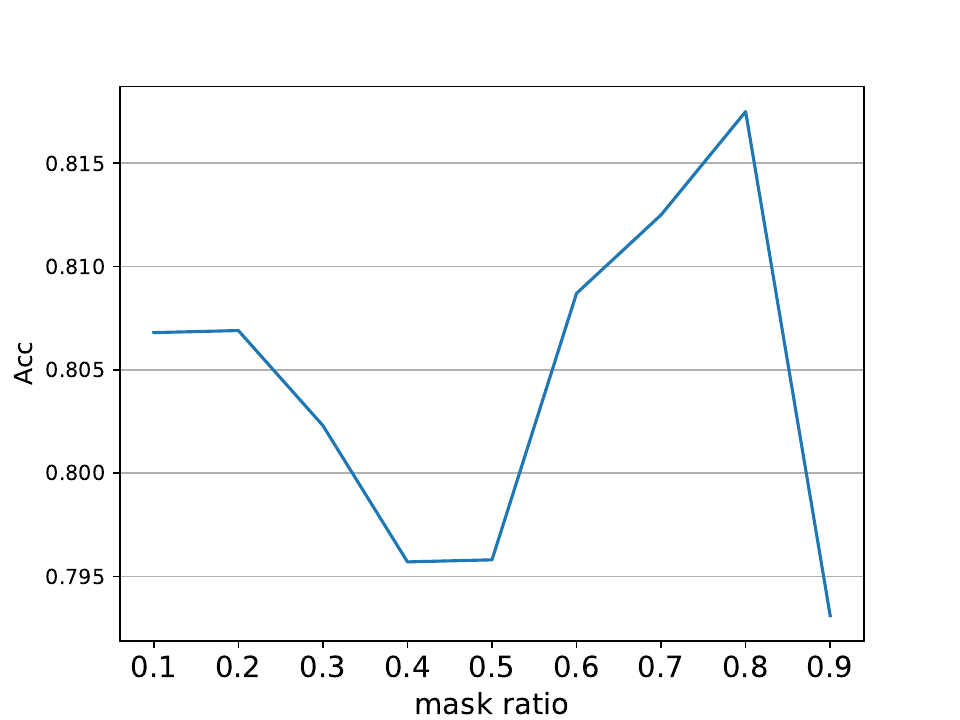}\label{fig:cox2 mask ratio}}
    \subfloat[ER\_MD]{\includegraphics[width=0.25\linewidth]{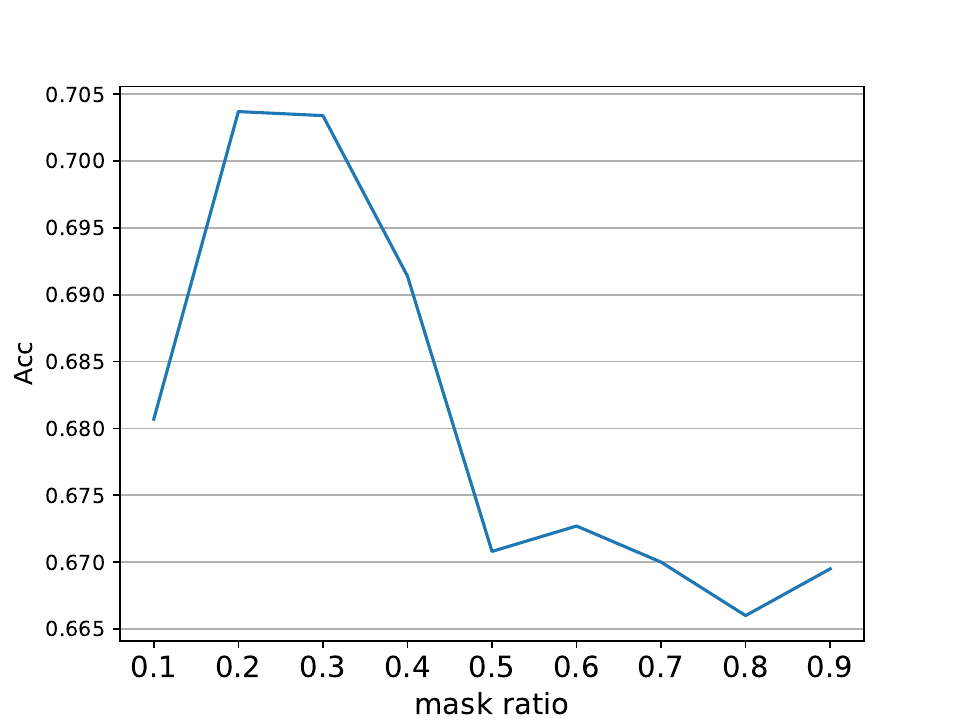}\label{fig:ermd mask ratio}}
    \quad
    \subfloat[NCI1]{\includegraphics[width=0.25\linewidth]{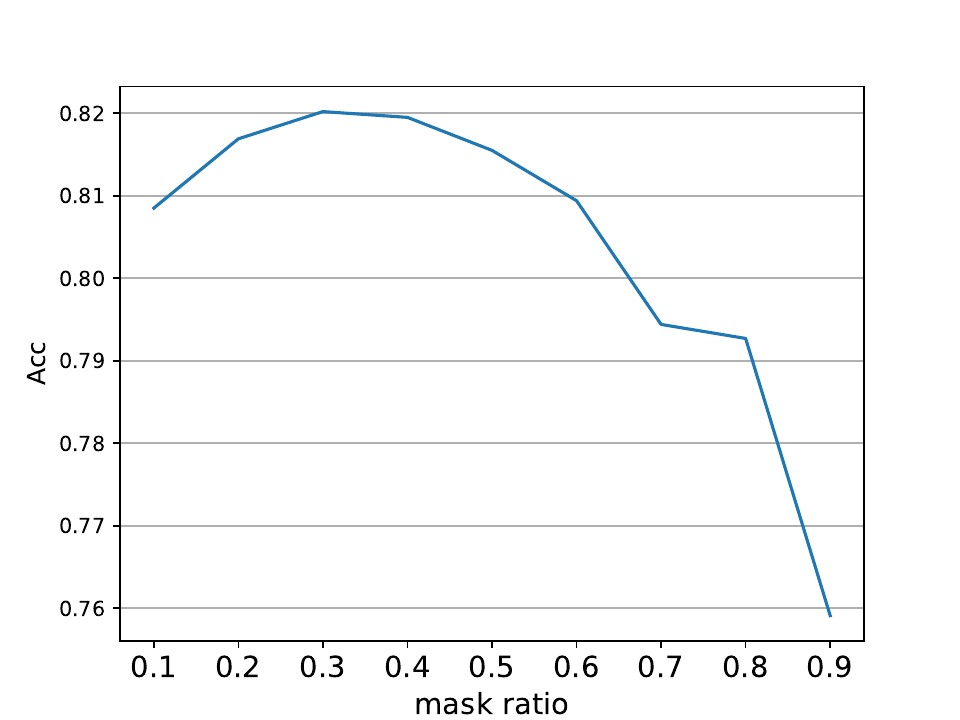}\label{fig:nci1 mask ratio}}
    \subfloat[MUTAG]{\includegraphics[width=0.25\linewidth]{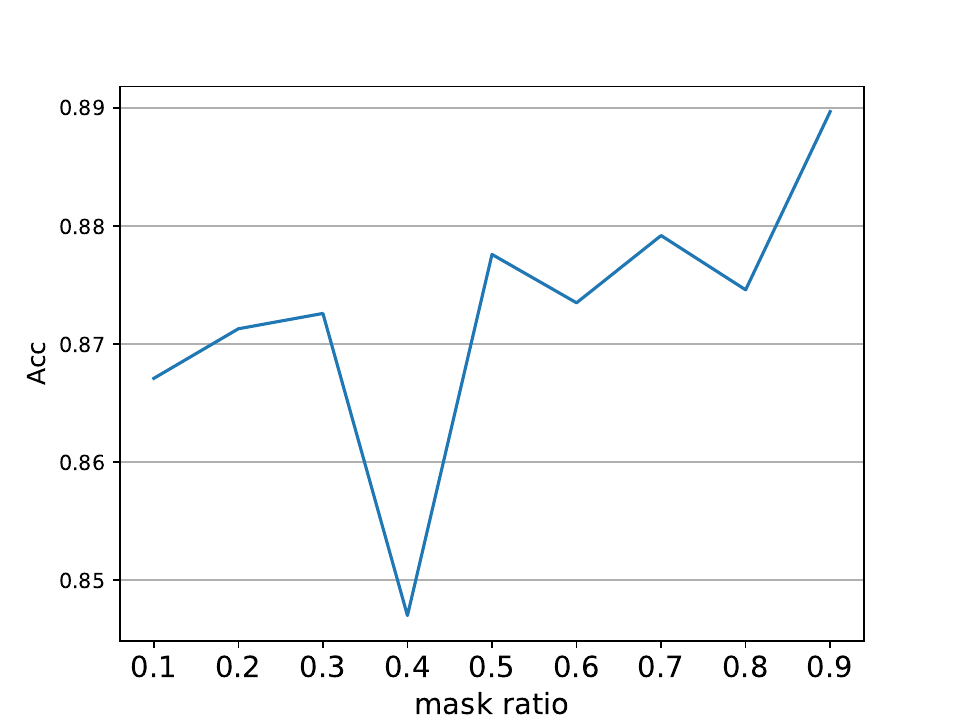}\label{fig:mutag mask ratio}}
    \subfloat[DHFR]{\includegraphics[width=0.25\linewidth]{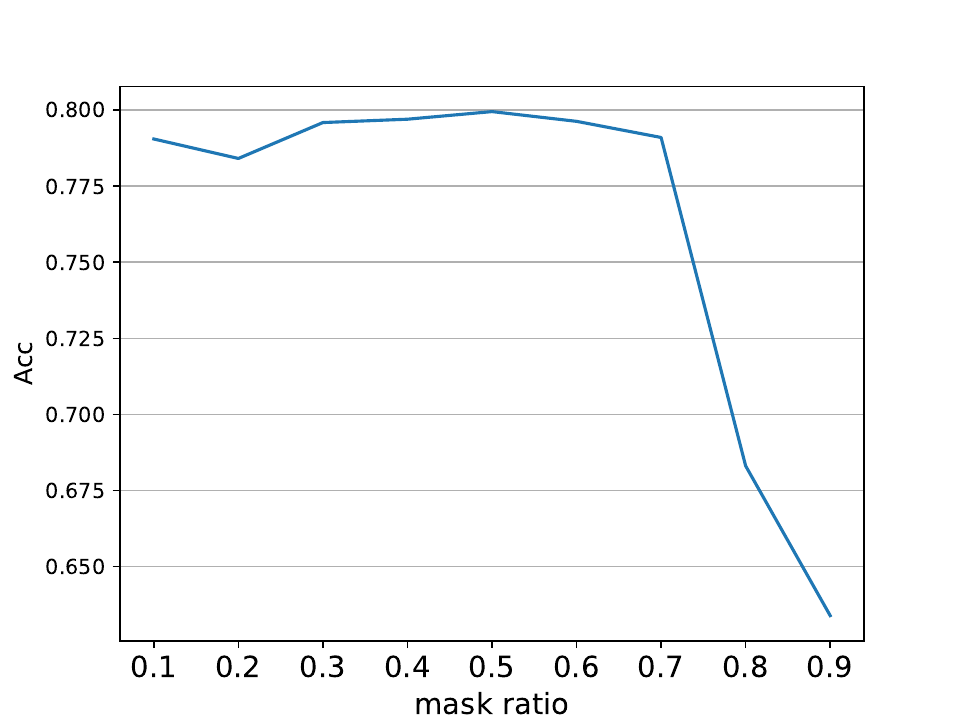}\label{fig:dhfr mask ratio}}
\caption{Mask ratios. The y axis is the accuracy and the x axis is the mask ratio. A larger accuracy indicates better performance.}
\label{fig:mask ratio}
\end{figure*}

\begin{figure*}[!t]
\centering
    \subfloat[PROTEINS]{\includegraphics[width=0.25\linewidth]{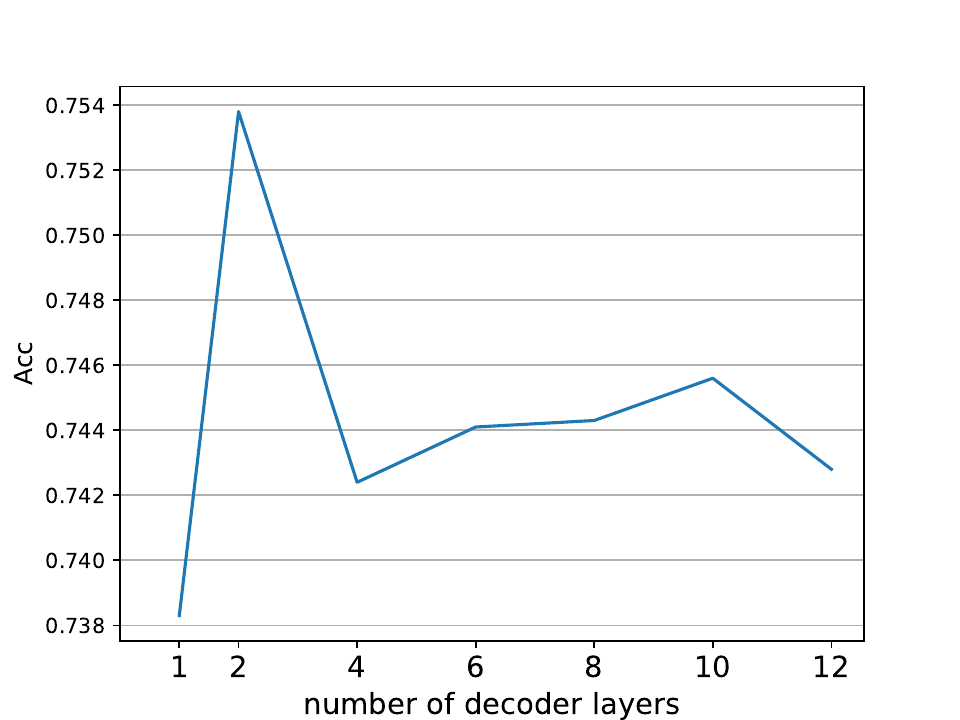}\label{fig:proteins n layers}}
    \subfloat[BZR]{\includegraphics[width=0.25\linewidth]{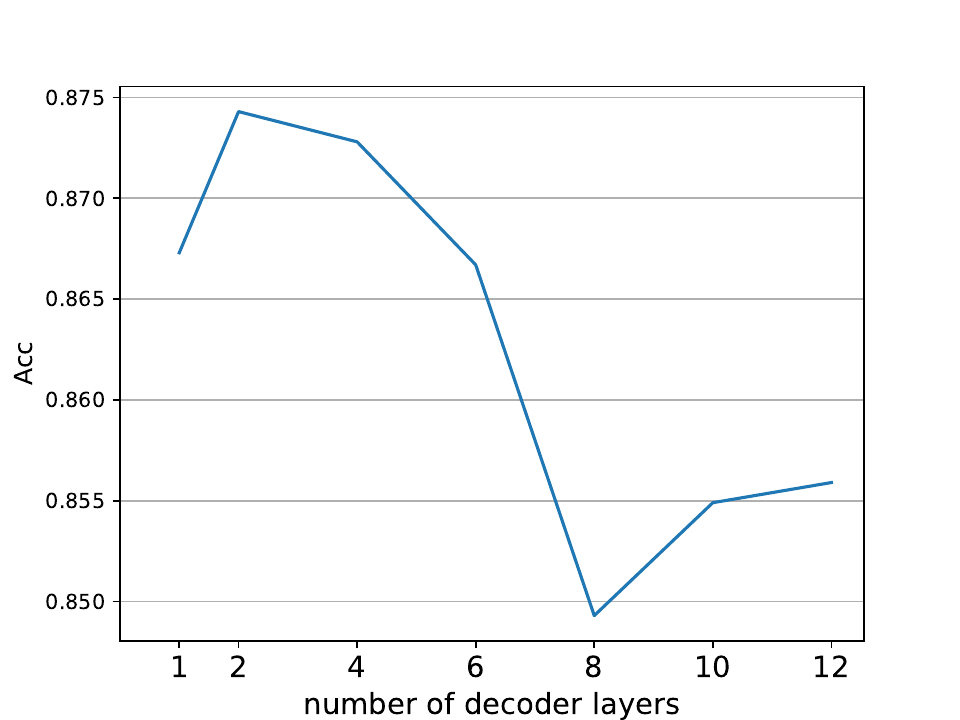}\label{fig:bzr n layers}}
    \subfloat[COX2]{\includegraphics[width=0.25\linewidth]{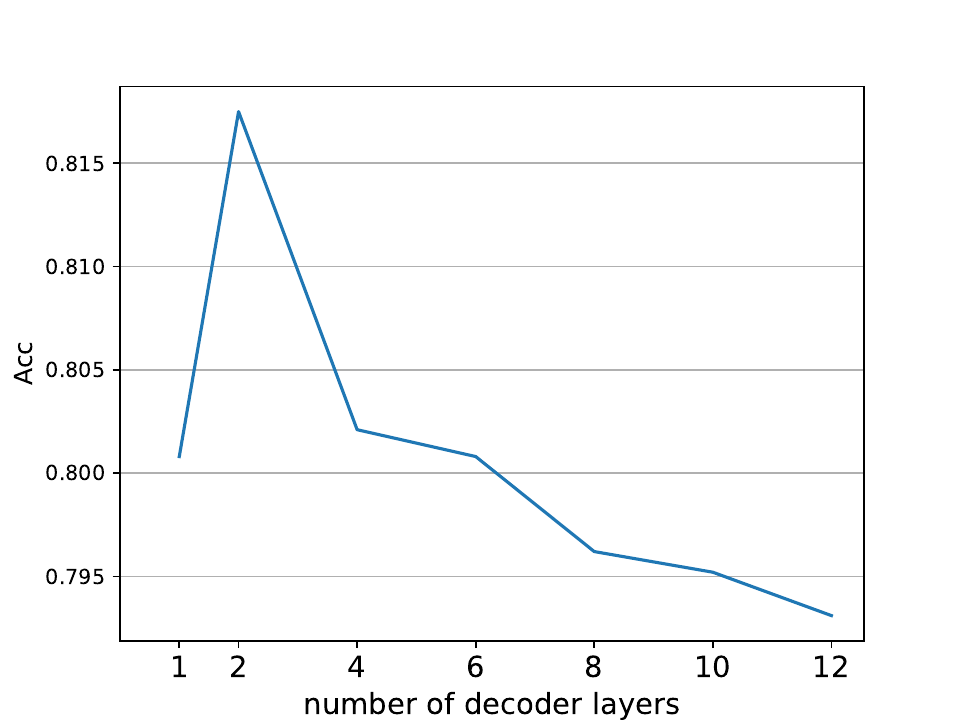}\label{fig:cox2 n layers}}
    \subfloat[ER\_MD]{\includegraphics[width=0.25\linewidth]{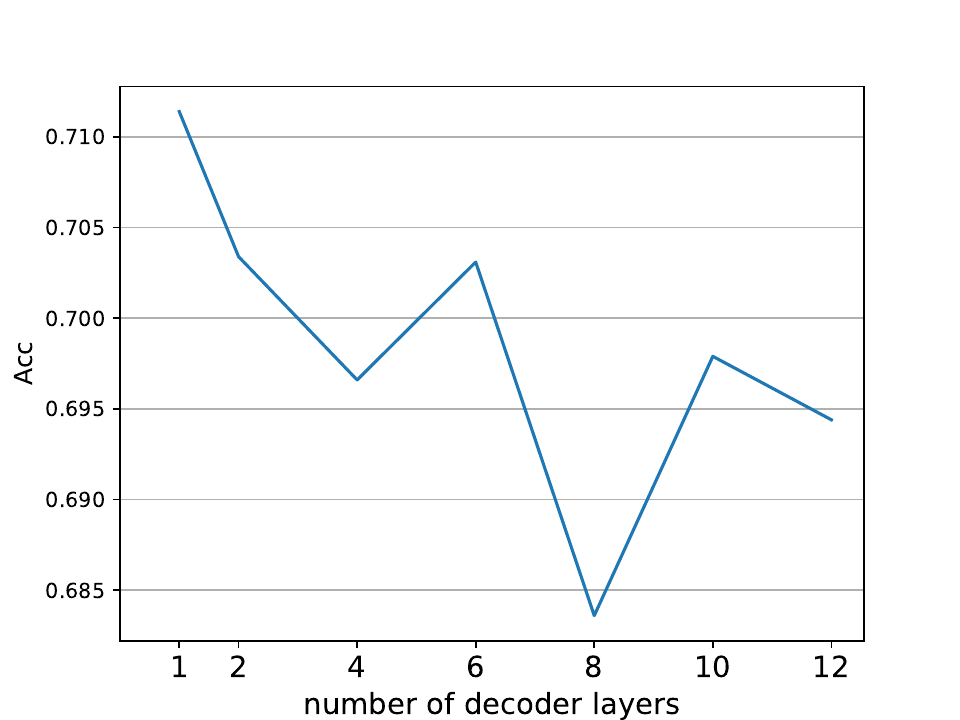}\label{fig:ermd n layers}}
    \quad
    \subfloat[NCI1]{\includegraphics[width=0.25\linewidth]{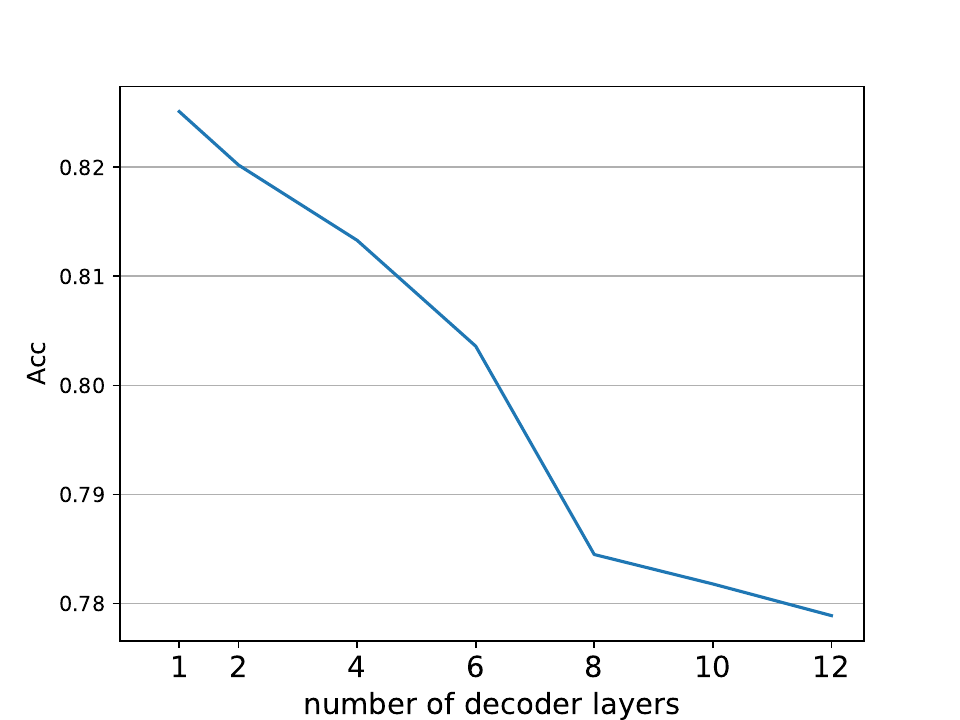}\label{fig:nci1 n layers}}
    \subfloat[MUTAG]{\includegraphics[width=0.25\linewidth]{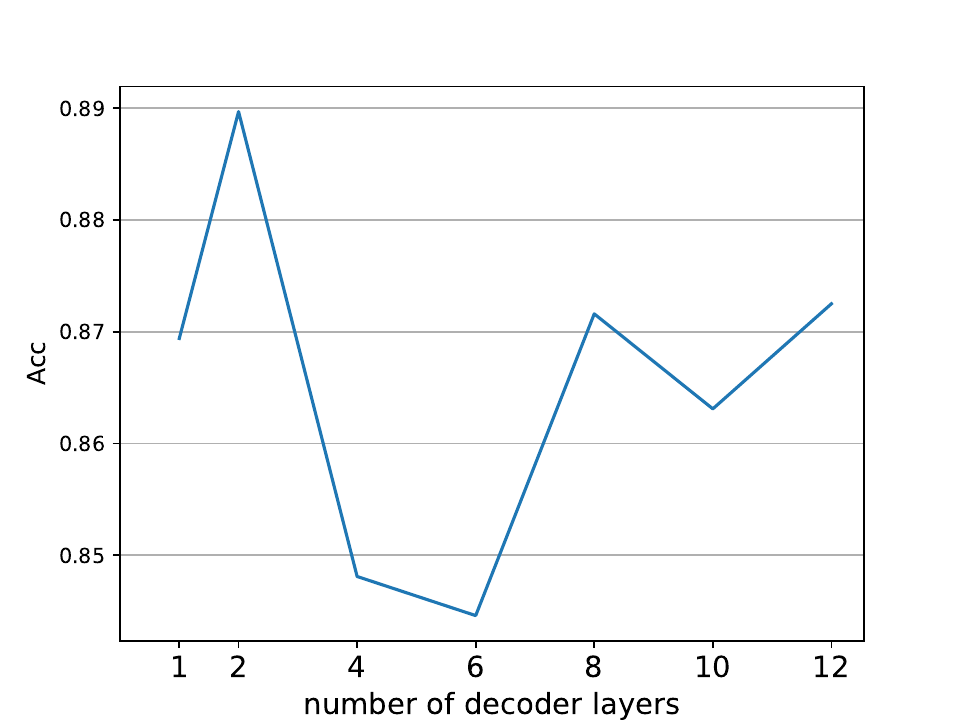}\label{fig:mutag n layers}}
    \subfloat[DHFR]{\includegraphics[width=0.25\linewidth]{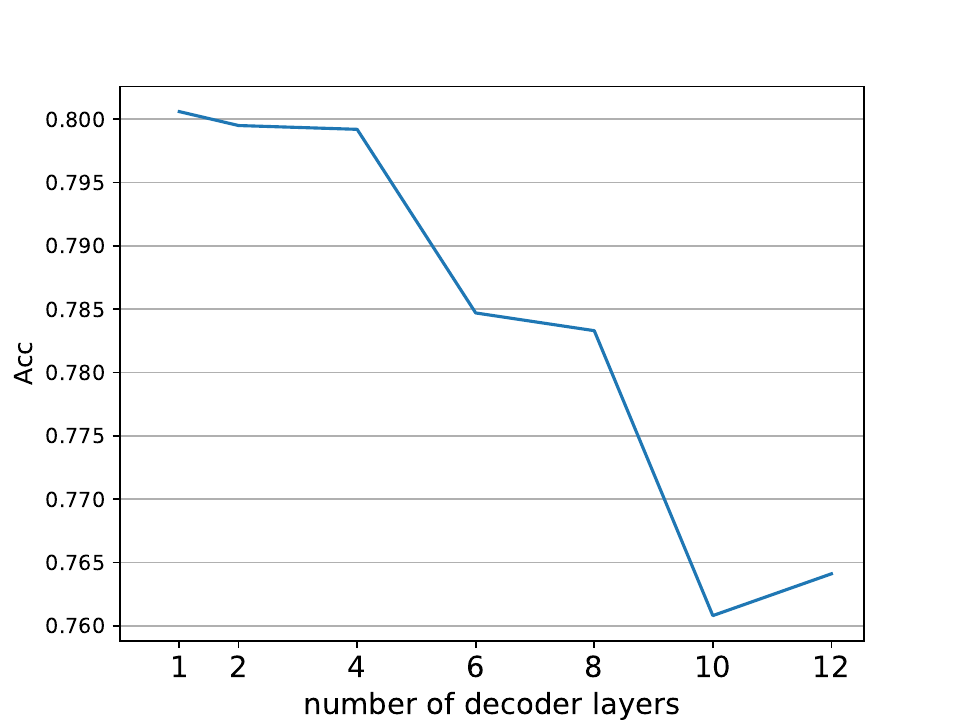}\label{fig:dhfr n layers}}
\caption{Number of decoder layers. The y axis is the accuracy and the x axis is the number of decoder layers. A larger accuracy indicates better performance.}
\label{fig:n layers}
\end{figure*}

Our GMAE shows slightly better performance than the state-of-the-art Graphormer. However, GMAE does not need a very tailored training scheme. The pre-training enables the model to learn knowledge of the inherent data structures and features. Such knowledge can be transferred to specific downstream tasks with a relatively quick learning process where a larger learning rate and fewer epochs will work.

\subsection{Analysis}
We show some basic properties of GMAE, including how much memory it saves compared with conventional transformers, and how it performs with different mask ratios and numbers of decoder layers.
\subsubsection{Memory Consumption}
We first compare the memory consumption of GMAE and Graphormer. For Graphormer, we set the number of layers to 12. For GMAE, the encoder is a 12-layer transformer and the decoder is a 2-layer transformer. We set the mask ratio in GMAE to 0.7. The memory consumption of the two models w.r.t. the number of nodes is shown in \autoref{fig:memory}. We can observe that, the memory usage of GMAE is significantly less than Graphormer. As the number of nodes increases, the memory saved by GMAE also increases. The space complexity for Graphormer is $O(n^{2})$, where $n$ is the number of nodes. With 0.7 mask ratio, the space complexity for GMAE is $O(0.49n^{2})$. Theoretically, GMAE can save around half of the memory in the self-attention module of the encoder. However, due to the extra memory consumption in other parts of the model, we can only observe a 20\% memory reduction at 110 nodes in \autoref{fig:memory}. But it still suggests that GMAE can be potentially used to train larger graphs and deeper graph transformers.

\subsubsection{Mask Ratio}
Next, we show how the mask ratio influences the performance of GMAE. We conduct experiments on the self-supervised graph classification scenario (using 7 benchmark datasets from TU dataset). For each dataset, we tune the mask ratio from 0.1 to 0.9 while keeping other hyper-parameters consistent with the ones used in \autoref{table:graph classification performance}. We report the downstream graph classification accuracy for the 7 datasets, which are all averaged over 5 runs. The results are shown in \autoref{fig:mask ratio}. Note that the y axis is the accuracy, so a larger value indicates a better performance.

Generally, the best mask ratio depends on the specific dataset. For example, COX2 and MUTAG prefer a large mask ratio at around 0.8 and 0.9 respectively; ER\_MD and NCI1 prefer a small mask ratio at around 0.3; PROTEINS, however, has two peak values at 0.4 and 0.7; BZR works well with all the mask ratios ranging from 0.2 to 0.7, while DHFR prefers mask ratios smaller than 0.7. In general, GMAE has a good overall performance when the mask ratio is around 0.4 or 0.7. In other words, GMAE can achieve very good performance when around half of the nodes are masked, which can lead to a significant memory reduction.

\subsubsection{Number of Decoder Layers}
At last, we show how the depth of the decoder influences the performance of GMAE. We follow the settings we used when comparing different mask ratios. However, this time we tune the number of decoder layers instead of the mask ratio. The results are shown in \autoref{fig:n layers}. We can observe that, a shallow decoder with 1 or 2 layers generally performs better than a deep decoder, which suggests that such an asymmetric encoder-decoder design works well for GMAE. It can not only maintain good performance but also help save memory.

\section{Conclusion}
In this paper, we present a novel self-supervised graph representation model named Graph Masked Autoencoders (GMAEs) based on transformers. It masks a large portion of the nodes in a graph and reconstructs the features of those masked nodes. It has state-of-the-art performance when evaluated under the commonly used self-supervised downstream evaluation protocols on both the graph classification task and the node classification task. When serving as a pre-training tool, it can help simplify the training process of graph transformers, while achieving comparable performance compared with training in an end-to-end manner. With the masking mechanism and asymmetric encoder-decoder design, the memory consumption of GMAE is largely reduced compared with conventional graph transformers, and can therefore be used to train large graphs and deep transformers.


%



\ifCLASSOPTIONcompsoc
  \section*{Acknowledgments}
\else
  \section*{Acknowledgment}
\fi

The work has been supported by Australian Research Council under grants DP220103717, DP200101374, LP170100891, and LE220100078.

\ifCLASSOPTIONcaptionsoff
  \newpage
\fi



\bibliographystyle{IEEEtran}
\bibliography{IEEEabrv,ref}
%



%

\begin{IEEEbiography}[{\includegraphics[width=1in,height=1.25in,clip,keepaspectratio]{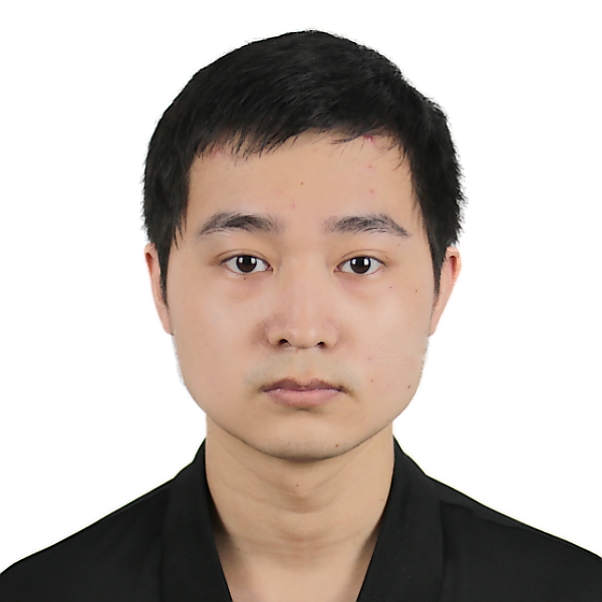}}]{Sixiao Zhang} is currently a PhD student of Data Science and Machine Intelligence (DSMI) Lab of Advanced Analytics Institute, University of Technology Sydney. He obtained his master's degree from Case Western Reserve University in the US, and obtained his bachelor's degree from University of Science and Technology of China. He is interested in the broad graph mining area and its downstream applications, including graph neural networks, recommender systems, adversarial robustness, etc. He has published papers in top-tier conferences and journals including KDD, WSDM, WWW, etc.
\end{IEEEbiography}

\begin{IEEEbiography}[{\includegraphics[width=1in,height=1.25in,clip,keepaspectratio]{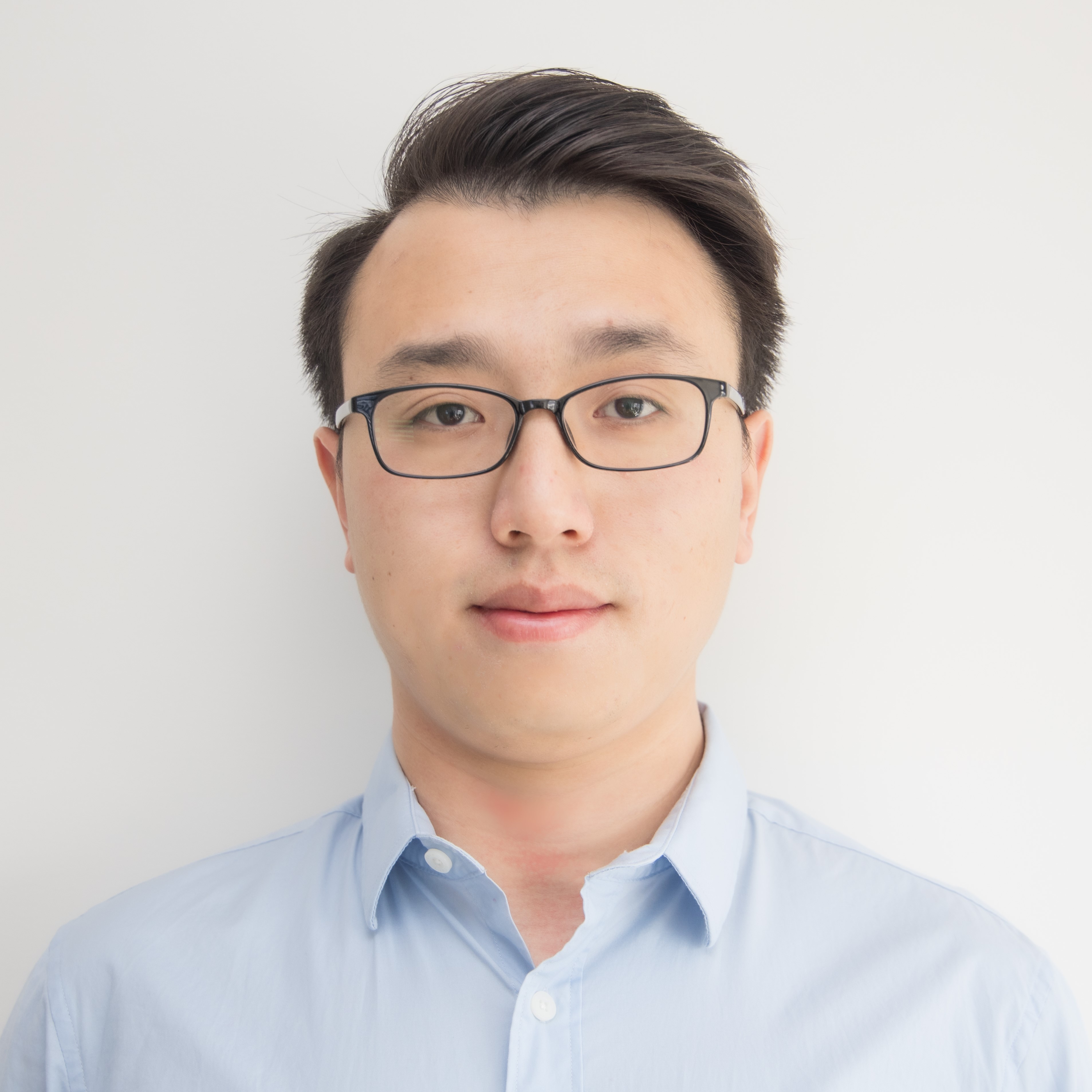}}]{Hongxu Chen} is a Data Scientist, now working as a Postdoctoral Research Fellow in School of Computer Science at University of Technology Sydney, Australia. He obtained his Ph.D. in Computer Science at The University of Queensland in 2020. His research interests mainly focus on data science in general and expand across multiple practical application scenarios, such as network science, data mining, recommendation systems and social network analytics. In particular, his research is focusing on learning representations for information networks and applying the learned network representations to solve real-world problems in complex networks such as biology, e-commerce and social networks, financial market and recommendations systems with heterogeneous information sources. He has published many peer-reviewed papers in top-tier high-quality international conferences and journals, such as SIGKDD, ICDE, ICDM, AAAI, IJCAI, TKDE. He also serves as a program committee member and reviewer in multiple international conferences, such as CIKM, ICDM, KDD, SIGIR, AAAI, PAKDD, WISE, and he also acts as an invited reviewer for multiple journals in his research fields, including Transactions on Knowledge and Data Engineering (TKDE), WWW Journal, VLDB Journal, IEEE Transactions on Systems, Man and Cybernetics: Systems, Journal of Complexity, ACM Transactions on Data Science, Journal of Computer Science and Technology.
\end{IEEEbiography}

\begin{IEEEbiography}[{\includegraphics[width=1in,height=1.25in,clip,keepaspectratio]{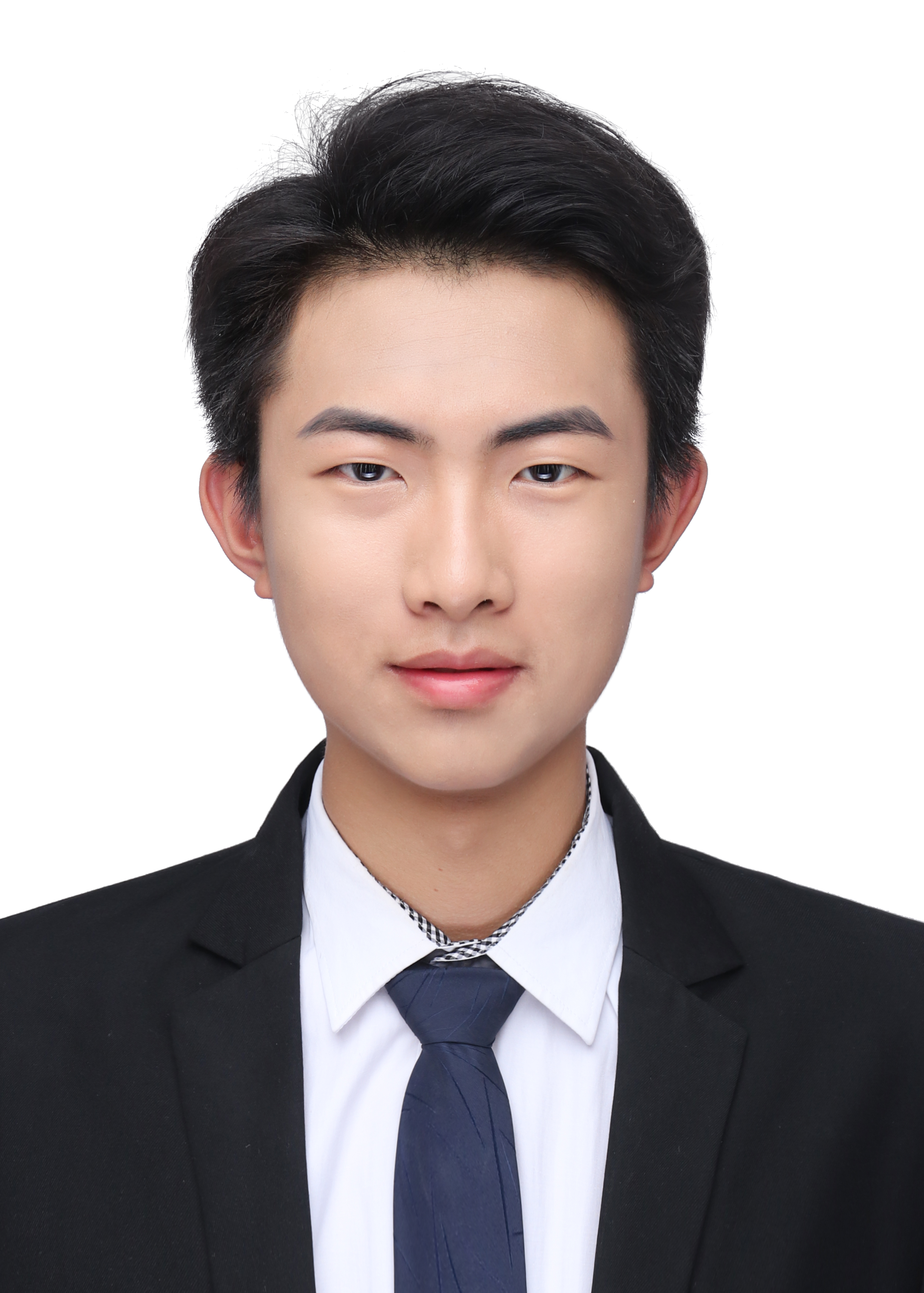}}]{Haoran Yang} is currently a Ph.D. candidate at Faculty of Engineering and Information Technology, University of Technology Sydney, who is supervised by Prof. Guandong Xu and Dr. Hongxu Chen. He obtained his bachelor's degree in computer science and technology at Nanjing University in 2020. His research interests include but not limited to graph representation learning and its applications, such as graph contrastive learning and recommendation systems. He has published several papers in top-tier conferences in his research areas, including WWW and ICDM. Haoran is also invited to serve as a reviewer of many international conferences and journals, such as KDD, IJCNN, and WWWJ.
\end{IEEEbiography}

\begin{IEEEbiography}[{\includegraphics[width=1in,height=1.25in,clip,keepaspectratio]{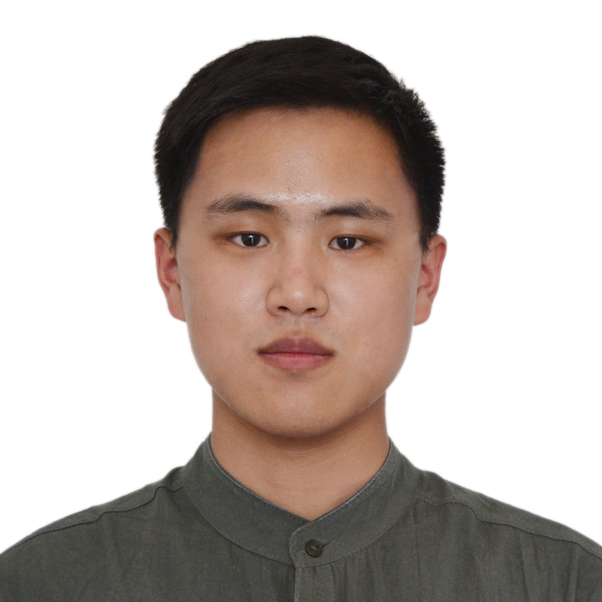}}]{Xiangguo Sun} works as a postdoctoral research fellow at the Chinese University of Hong Kong and works with Prof. Hong Cheng. He is also a visiting researcher at the Research Institute of Artificial Intelligence in Zhejiang Lab hosted by Dr. Hongyang Chen. He received his Ph.D. from Southeast University under the supervision of Prof. Bo Liu (in Jan 2022, Nanjing). During his Ph.D. study, he worked as a research intern in Microsoft Research Asia with Dr. Hang Dong and Bo Qiao (from Sep 2021 to Feb 2022, Beijing), and visited the University of Queensland as a research scholar hosted by Prof. Hongzhi Yin (from Sep 2019 to Sep 2021, Australia). His research interests include social media analytics, and hypergraph learning. His work has been published in some of the most prestigious venues such as SIGKDD, TKDE, TNNLS, The Web Conference (WWW), WSDM, CIKM and et al. 
\end{IEEEbiography}

\begin{IEEEbiography}[{\includegraphics[width=1in,height=1.25in,clip,keepaspectratio]{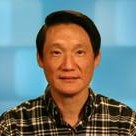}}]{Philip S. Yu} is a Distinguished Professor in Computer Science at the University of Illinois at Chicago and also holds the Wexler Chair in Information and Technology. His research interests
include data mining, privacy preserving publishing and mining, data streams, database systems, Internet applications and technologies, multimedia systems, parallel and distributed processing, and performance modeling. Dr. Yu has published more than 970 papers in refereed journals
and conferences with more than 143,604 citations and an H-index of 1215. He holds or has applied for more than 300 US patents. He is a Fellow of the ACM and of the IEEE. He is associate editors of ACM Transactions on the Internet Technology and ACM Transactions on Knowledge Discovery from Data. He is on the steering committee of IEEE Conference on Data Mining and was a member of the IEEE Data Engineering steering committee. He was the Editor-in-Chief of IEEE Transactions on Knowledge and Data Engineering (2001-2004), an editor, advisory board member and also a guest co-editor of the special issue on mining of databases. He had also served as an associate editor of Knowledge and Information Systems. He also received an IEEE Region 1 Award for ”promoting and perpetuating numerous new electrical engineering concepts” in 1999. He had received several UIC honors, including Research of the Year at 2013 and UI Faculty Scholar at 2014. He also received many IBM honors including 2 IBM Outstanding Innovation Awards, an Outstanding Technical Achievement Award, 2 Research Division Awards and the 94th plateau of Invention Achievement Awards. He was an IBM Master Inventor. Dr. Yu received the B.S. Degree in E.E. from National Taiwan University, the M.S. and Ph.D. degrees in E.E. from Stanford University, and the M.B.A. degree from New York University.
\end{IEEEbiography}

\begin{IEEEbiography}[{\includegraphics[width=1in,height=1.25in,clip,keepaspectratio]{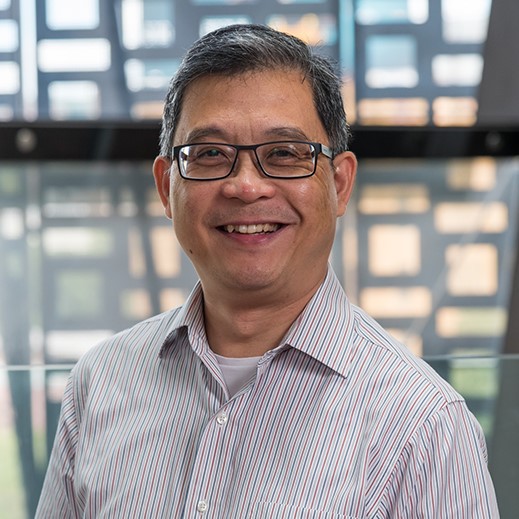}}]{Guandong Xu} is a professor in the School of Computer Science and Data Science Institute at UTS and an award-winning researcher working in the fields of data mining, machine learning, social computing and other associated fields. He is Director of the UTS-Providence Smart Future Research Centre, which targets research and innovation in disruptive technology to drive sustainability. He also heads the Data Science and Machine Intelligence Lab, which is dedicated to research excellence and industry innovation across academia and industry, aligning with the UTS research priority areas in data science and artificial intelligence. Guandong has had more than 220 papers published in the fields of Data Science and Data Analytics, Recommender Systems, Text Mining, Predictive Analytics, User Behaviour Modelling, and Social Computing in international journals and conference proceedings in recent years, with increasing citations from academia. He has shown strong academic leadership in various professional activities. He is the founding Editor-in-Chief of Humancentric Intelligent System Journal, the Assistant Editor-in-Chief of World Wide Web Journal, as well as the founding Steering Committee Chair of the International Conference of Behavioural and Social Computing Conference.
\end{IEEEbiography}




\end{document}